\documentclass[10pt,journal,compsoc]{IEEEtran}
\usepackage{cite}
\usepackage[pdftex]{graphicx}
\usepackage{amsmath}
\usepackage{array}
\usepackage{multirow}
\usepackage{xcolor}
\usepackage{bbm}
\usepackage{amsmath}
\usepackage{amssymb}
\usepackage{enumitem}
\usepackage{subcaption}
\usepackage{bbding}
\usepackage{pifont}

\definecolor{Olive_Green}{rgb}{0.0, 0.55, 0.0}

\usepackage[pagebackref=false,breaklinks=true,bookmarks=false]{hyperref}

\newcolumntype{P}[1]{>{\centering\arraybackslash}p{#1}}

\usepackage{url}
\usepackage{booktabs}       
\usepackage{amsfonts}       
\usepackage{nicefrac}       
\usepackage{microtype}      
\usepackage{xcolor}    
\usepackage{multirow}
\usepackage{adjustbox}
\usepackage{amssymb}
\usepackage{pifont}
\usepackage{amsmath}
\newcommand{\et}{\textit{et al.\ }}

\begin{document}
\title{SimBase: A Simple Baseline\\for Temporal Video Grounding}

\author{
Peijun Bao,
\and
Alex C. Kot,~\IEEEmembership{Life~Fellow,~IEEE}
\IEEEcompsocitemizethanks{
\IEEEcompsocthanksitem
Technical report.
\IEEEcompsocthanksitem 
Peijun Bao,  and Alex C. Kot are with the School of Electrical and Electronic Engineering, Nanyang Technological University, Singapore.
%\protect\\
%
%\IEEEcompsocthanksitem 
\IEEEcompsocthanksitem{
Peijun Bao is the corresponding author. E-mail: peijun001@e.ntu.edu.sg
}
}
}

% The paper headers
%\markboth{Journal of \LaTeX\ Class Files,~Vol.~14, No.~8, August~2024}%
%{Shell \MakeLowercase{\textit{et al.}}: Bare Demo of IEEEtran.cls for Computer Society Journals}
\markboth{Technical Report}%
{Shell \MakeLowercase{\textit{et al.}}: Bare Demo of IEEEtran.cls for Computer Society Journals}

\IEEEtitleabstractindextext{%
\begin{abstract}
This paper presents SimBase, a simple yet effective baseline for temporal video grounding.
While recent advances in temporal grounding have led to impressive performance, they have also driven network architectures toward greater complexity, with a range of methods to (1) capture temporal relationships and (2) achieve effective multimodal fusion.
In contrast, this paper explores the question: \textit{How effective can a simplified approach be}?
To investigate, we design SimBase, a network that leverages lightweight, one-dimensional temporal convolutional layers instead of complex temporal structures.
For cross-modal interaction, SimBase only employs an element-wise product instead of intricate multimodal fusion.
Remarkably, SimBase achieves state-of-the-art results on two large-scale datasets.
As a simple yet powerful baseline, we hope SimBase will spark new ideas and streamline future evaluations in temporal video grounding.
\end{abstract}
}

\maketitle

\IEEEdisplaynontitleabstractindextext

\ifCLASSOPTIONpeerreview
\begin{center} \bfseries EDICS Category: 3-BBND \end{center}
\fi

\IEEEpeerreviewmaketitle

\begin{figure}[t!]
\centering
\includegraphics[width=0.8\linewidth]{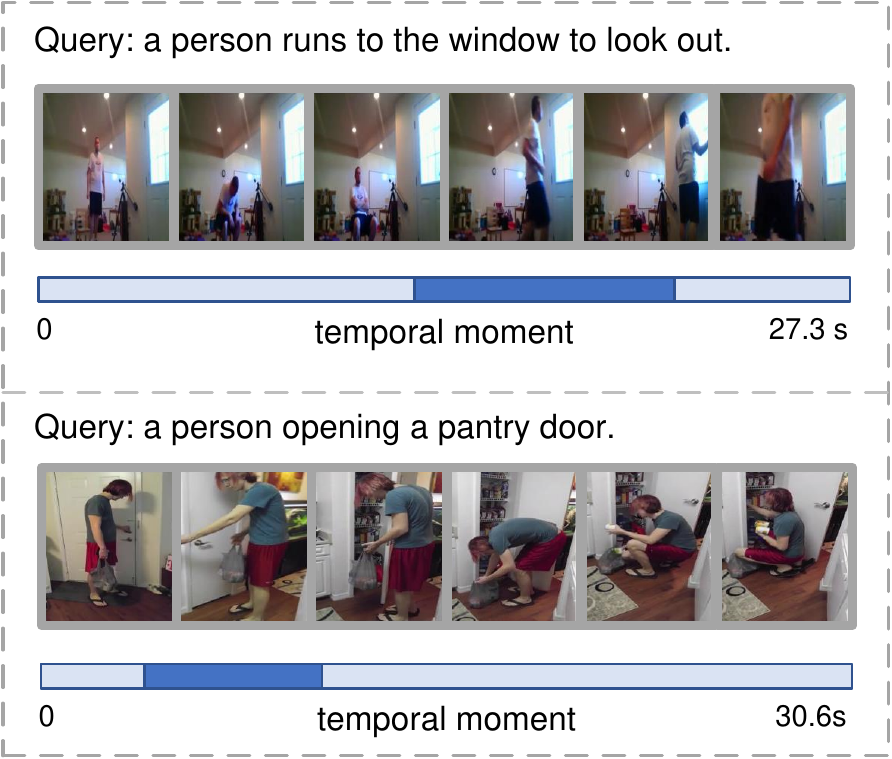}
\caption{
Examples of temporal video grounding.
The goal of temporal video grounding is to localize the temporal moment in an untrimmed video based on a given sentence query.
}\label{fig_intro}
\end{figure}

\IEEEraisesectionheading{\section{Introduction}\label{sec:introduction}}
\IEEEPARstart{G}{iven} a natural language query and an untrimmed video, as shown in Fig~\ref{fig_intro}, the task of temporal video grounding~\cite{tall,dense_cap}  aims to predict the start and end time points of the video moment described by the sentence.
It is an important yet challenging task with a wide spectrum of applications in video understanding and analysis~\cite{Qi2021SemanticsAwareSB,Sreenu2019IntelligentVS,bao2024e3m,Zhu2021DSNetAF}.

Recent years have witnessed remarkable progress in temporal video grounding, driven by deep learning techniques~\cite{Liu2023TowardsBA,Moon2023QueryD,Bao2021DenseEG,Yan2023UnLocAU,Mun2020LocalGlobalVI,bao2024local,Bao2022LearningSI,bao2024omnipotent,Zhang2019CrossModalIN}.
Despite these advancements, the network architectures for video grounding have progressively become more complex, often involving intricate designs and a large number of parameters. 
Most of these designs focus on two primary objectives:
\begin{itemize}
\item
\textbf{Facilitating cross-modal fusion:} These approaches aim to integrate visual and textual modalities at different levels of granularity for precise grounding. 
Existing methods often devise variants of cross-attention mechanisms to effectively combine video frames and textual cues~\cite{Liu2023TowardsBA,Moon2023QueryD,Lee2023BAMDETRBM,Yan2023UnLocAU,Mun2020LocalGlobalVI,Zhang2019CrossModalIN}. enabling models to capture fine-grained interactions between modalities.
\item \textbf{Modeling temporal relationships:} Understanding the temporal dynamics within videos is essential. Techniques often leverage graph neural networks~\cite{Zhang2018MANMA} and various self-attention mechanisms~\cite{Liu2023TowardsBA,Moon2023QueryD,Yan2023UnLocAU,Bao2021DenseEG} to model temporal dependencies across different video frames or temporal proposals.
\end{itemize}
While these sophisticated methods lead to performance improvements on several benchmarks, they also introduce challenges in algorithm analysis and comparison.
For instance,  the increasing complexity makes it difficult to pinpoint which components contribute most to the performance gains.

This work addresses this issue by asking an opposite question: \textit{How effective can a simplified approach be}?
To explore this, we introduce SimBase, a simple yet powerful baseline for video grounding. 
SimBase leverages lightweight convolutional layers, relying on simple one-dimensional convolution instead of complex temporal modeling. 
For cross-modal interaction, it employs the basic element-wise product operations, avoiding intricate attention designs to fuse visual and textual modalities.

Surprisingly, SimBase achieves strong results on two large-scale benchmarks, notably  surpassing state-of-the-art method by a significant margin on the Charades-STA benchmark.
As a straightforward yet effective baseline, we hope SimBase will inspire innovative approaches and streamline the evaluation of new ideas for video grounding.

Our contributions can be summarized as:
\begin{itemize}
\item
We present SimBase, a simple baseline for temporal video grounding that primarily relies on temporal convolution for temporal modeling and element-wise product for multimodal fusion. 
\item
Experiments demonstrate that SimBase achieves state-of-the-art results on two widely used large-scale datasets. 
\end{itemize}

\begin{figure*}[t!]
\centering
%\fbox{\rule{0pt}{2.5in} \rule{\linewidth}{0pt}}
\includegraphics[width=1.0\linewidth]{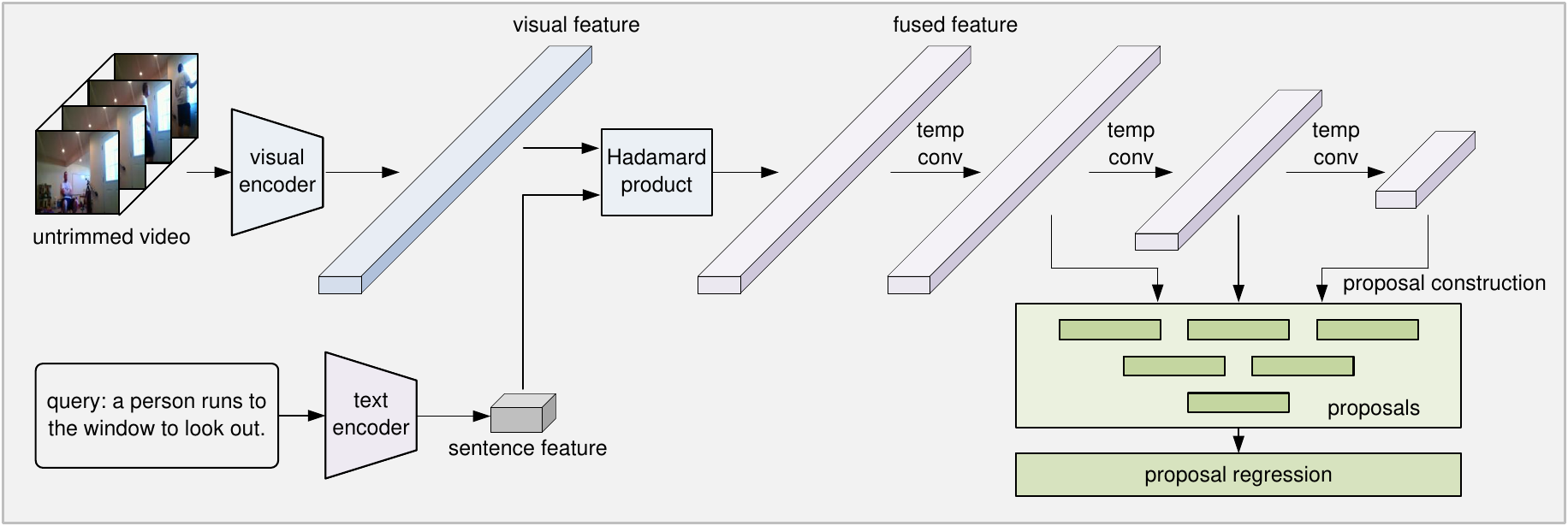}
%\vspace{-0.1cm}
\caption{
Overview of Simple Baseline (SimBase) for temporal video grounding.
SimBase leverages lightweight, one-dimensional temporal convolutional layers instead of complex temporal structures.
For cross-modal interaction, SimBase exploits Hadamard product rather than complex interaction of language and video.
}\label{fig_method}
\end{figure*}

\section{SimBase}
\subsection{Overview}
As illustrated in Fig~\ref{fig_method}, instead of complex temporal modeling, Simple Baseline (SimBase) for temporal video grounding leverages lightweight one-dimensional convolutional layers.
For cross-modal interaction, it uses straightforward element-wise product operations, avoiding the intricate attention mechanisms typically used to fuse visual and textual modalities.
Despite its simplicity, SimBase achieves state-of-the-art results on two large-scale datasets.

\subsection{Temporal Interaction and Multimodal Fusion}
We use the visual feature extractor of the CLIP model to extract each frame as a visual feature, denoted by $v_1, \ldots, v_L$.
We primarily use the simplest temporal convolution layer to enable the temporal interaction between different frames. 
A series of temporal convolution layers followed by ReLU are applied to the video frames to capture the temporal relationships among different frames.

Subsequently, we use the CLIP model to extract textual features from the sentence query $S$, represented as $s \in \mathbb{R}^N$.
We fuse the textual and visual features in the simplest way, namely using the Hadamard product, as follows:
\begin{equation}\label{eq_fusion}
\begin{aligned}
m_i &=  v_i \odot s
\end{aligned}
\end{equation}
where $\odot$ denotes Hadamard product.

After the fusion, we obtain the fused feature as $m^{(1)} \in \mathbb{R}^{L^{(1)} \times d}$.
A temporal convolution layer with a stride $s$, followed by batch normalization and ReLU layers, are applied to $m^{(1)}$ to facilitate interaction.
Consequently, the temporal resolution of the fused feature is reduced by a factor of $s$, yielding $m^{(2)} \in \mathbb{R}^{L^{(2)} \times d}$, where $L^{(2)}=L^{(1)}/s$.
This process is repeated $P-1$ times until $L^{P}=1$, thereby constructing a feature pyramid map.

\begin{table*}[t!]
\centering
\caption{
Comparisons with state-of-the-art methods on Charades-STA and ActivityNet Captions.
}
\vspace{-0.15cm}
\begin{tabular}{lcccccccccc}
\toprule
\multirow{2}{*}{Method} 
&\multirow{2}{*}{Venue} 
&\multirow{2}{*}{Vision enc.}
&\multicolumn{4}{c}{Charades-STA} 
&\multicolumn{4}{c}{ActivityNet Captions}  
\\ \cmidrule(lr){4-7} \cmidrule(lr){8-11} 
&&
&R@0.3   &R@0.5   &R@0.7 &mIoU
&R@0.3   &R@0.5   &R@0.7 &mIoU
\\ 
\midrule
CTRL~\cite{tall} 
& ICCV 2017
&\multirow{3}{*}{VGG/C3D}
& $-$   & 21.42 & 7.15 & $-$
& 28.70 & 14.00 & $-$ & $-$
\\
TripNet~\cite{Hahn2019TrippingTT} 
& BMVC 2020
&
&51.33  &36.61 &14.50 & $-$
&48.42  &32.19 &13.93 & $-$
\\
2D-TAN~\cite{Zhang2020Learning2T} 
& AAAI 2020
&
& $-$ &39.70 &23.31 & $-$
&59.45 &44.51 &26.54 & $-$
\\
\midrule
MAN~\cite{Zhang2018MANMA} 
&CVPR 2019
&\multirow{4}{*}{I3D/C3D}
& $-$  &46.53  &22.72 & $-$
& $-$  & $-$  & $-$ & $-$
\\
PfTML-GA\cite{RodriguezOpazo2019ProposalfreeTM} 
&WACV 2020
&
&67.53  &52.02  &33.74 & $-$
&51.28  &33.04  &19.26 & $-$
\\
SCDM~\cite{Yuan2019SemanticCD} 
&NeuRIPS 2019
&
& $-$   &54.44  &33.43 & $-$
&54.80  &36.75  &19.86 & $-$
\\
\midrule
QD-DETR~\cite{Moon2023QueryD} 
&CVPR 2023
&\multirow{5}{*}{CLIP}
& $-$  &56.89 &32.50 & $-$
& $-$  & $-$  & $-$ & $-$
\\
MESM~\cite{Liu2023TowardsBA} 
&AAAI 2024
& 
& $-$ & 61.24 & 38.04  & $-$
& $-$ & $-$ & $-$ & $-$
\\
BAM-DETR~\cite{Lee2023BAMDETRBM} 
&ECCV 2024
& 
&72.93 &59.95 &39.38 &52.33
& $-$ & $-$ & $-$ & $-$
\\
UnLoc~\cite{Yan2023UnLocAU} 
&ICCV 2023
&
& $-$ & 60.80 & 38.40 & $-$
& $-$ & 48.00 & 30.20 & $-$
\\
\textbf{SimBase (Ours)} 
& $-$
&
&\textbf{77.77} &\textbf{66.48} &\textbf{44.01} & \textbf{56.15}
&\textbf{63.98} &\textbf{49.35} &\textbf{30.48} & \textbf{47.07}
\\
\bottomrule
\end{tabular}
\label{table_sota}
\end{table*}

\subsection{Proposal Construction}
We represent each proposal as $(c, w)$, where $c$ and $w$ denote the center and the width of the proposal, respectively.
Let the temporal span in the $p$-th layer of the pyramid feature map $m^{(p)} \in \mathbb{R}^{L^{(p)} \times d}$ be $T^{(p)}$ in the original video. We then define a list of proposals $(c^p_i, w^p_i)$ on it.
The proposals are constructed by expanding the temporal span of features with different scale ratios $r_1, \ldots, r_K$, where $K$ is the number of scale ratios.
To ensure proper division, we select the original $L^{1}$ and stride $s$ such that $L^{1} = s^P$.
For each proposal $(c^p_i, w^p_i)$, we predict its Intersection-over-Union (IoU) value $IoU_i^p$ by applying a convolutional layer followed by a sigmoid activation function on the fused feature map.

\subsection{Proposal Regression}
To refine the temporal boundaries derived from a fixed proposal with center $\widehat{c}$ and width $\widehat{w}$, we further regress the offsets of the temporal boundaries to adjust their original values.
For each proposal, we consider two offsets for regression: the center offset $\delta_c$ and the width offset $\delta_w$.
We apply a convolutional layer followed by a sigmoid activation function to the fused feature map to predict $\delta_c$ and $\delta_w$.

The original center $\widehat{c}$ and width $\widehat{w}$ of the temporal proposal are then adjusted as follows:
\begin{equation}
\begin{aligned}
c &= \widehat{c} + \alpha \cdot w \cdot \delta_c \\
w &= \widehat{w} \cdot \exp \left( \beta \cdot \delta_w \right)
\end{aligned}
\end{equation}
Here, $\alpha$ and $\beta$ are linear factors.

\subsection{Training loss}
Let the ground truth of temporal boundary is $\overline{b} = (\overline{c}, \overline{w})$.
For each proposal, we first compute its overlap with the ground truth temporal boundary as
$\overline{\theta}$.
We determine its offset from the true boundary as follows:
\begin{equation}
\begin{aligned}
\overline{\delta}_c &= \frac{\overline{c} - \widehat{c}}{\alpha \cdot w}\\
\overline{\delta}_w &= \frac{1}{\beta} \log \left( \overline{w} / \widehat{w} \right)
\end{aligned}
\end{equation}

Subsequently, we align the model prediction to the ground truth by using IoU loss $\mathcal{L}_{\text{iou}}$ and regression loss $\mathcal{L}_{\text{reg}}$.
Specifically, the prediction of the IoU value is enforced to be consistent with the ground truth through cross-entropy loss, formulated as:
\begin{equation}
\mathcal{L}_{\text{iou}} =
- \overline{\theta} \log \theta
- \left( 1- \overline{\theta} \right) \log \left( 1- \theta \right)
\end{equation}

Moreover, we apply a smooth L1 loss~\cite{Girshick2015FastR} to enforce the model to predict the offset of the proposal as close as possible to the target offset:
\begin{equation}
\mathcal{L}_{\text{reg}}  =
\mathcal{L}_{\text{smooth}}\left(\overline{\delta}_c -  \delta_c\right)
+ \mathcal{L}_{\text{smooth}}\left(\overline{\delta}_w -  \delta_w\right)
\end{equation}
where $\delta_c$ and $\delta_w$ is the model's prediction of the offset, and $SL$ denote the Smooth L1 loss.
The final loss is the combination of $\mathcal{L}_{\text{IoU}}$ and $\mathcal{L}_{\text{refine}}$, formulated as
\begin{equation}
\mathcal{L}_{\text{SimBase}}  =
\mathcal{L}_{\text{iou}} + \lambda \mathcal{L}_{\text{reg}}
\end{equation}
where $\lambda$ is a balancing hyperparameter.

\section{Experiments}
\subsection{Datasets}
We validate the performance of the proposed methods against the state-of-the-art methods on two large-scale datasets: Charades-STA~\cite{tall} and ActivityNet Captions~\cite{dense_cap}.
\noindent
\textbf{Charades-STA} contains 9,848 videos of daily indoor activities.
The average length of a sentence query is 8.6 words, and the average duration of the video is 29.8 seconds.
It is originally designed for action recognition / localization~\cite{Sigurdsson2016HollywoodIH}, and later extended by Gao \et~\cite{tall} with language descriptions for video temporal grounding.

\noindent
\textbf{ActivityNet Captions} consists of 19,290 untrimmed videos, whose contents are diverse and open. 
The average duration of the video is 117.74 seconds and the average length of the description is 13.16 words.
There are 2.4 annotated moments with a duration of 8.2 seconds in each video.

\subsection{Evaluation Metrics}
Following previous works, we adopt the evaluation metric `R@m' for video moment retrieval to evaluate the performance of our method. 
Specifically, we calculate the Intersection over Union (IoU) between the retrieved temporal moment and the ground truth. 
Then `R@m' is defined as the percentage of language queries having correct moment retrieval results, where a moment retrieval is correct if its IoU is larger than $m$.
As in previous works, we report the results with $m=\{0.3,0.5,0.7\}$ on both datasets.

\subsection{Implementation Details}
The number of video snippets $N$ is set to $128$ and $256$ on Charades-STA and ActivityNet Captions respectively.
We use the CLIP~\cite{Radford2021LearningTV} model to extract visual and textual features.
The feature channel $d$ is set to $768$ and the kernel size of the temporal convolution is set to $3$.
The hyperparameter $\lambda$ is set to $0.0025$.
The linear factors $\alpha$ and $\beta$ are set to 0.3 for Charades STA, and to 0.4 and 0.2 for ActivityNet Captions, respectively.
We train our model using the Adam optimizer~\cite{kingma2014adam} with a batch size of $32$ and a learning rate of $0.0008$.
The training epoch number is set to $20$.

\subsection{Performance Comparisons}
Table~\ref{table_sota} presents a comparison between SimBase and a list of state-of-the-art methods. 
On the Charades STA dataset, SimBase significantly outperforms recent methods, including QD-DETR~\cite{Moon2023QueryD}, MESM~\cite{Liu2023TowardsBA}, BAM-DETR~\cite{Lee2023BAMDETRBM}, and UnLoc~\cite{Yan2023UnLocAU}. 
For instance, SimBase exceeds MESM by over 5 points in R@0.7 and surpasses BAM-DETR by approximately 4 points in mIoU.
While UnLoc and BAM-DETR employ complex attention-based architectures for temporal modeling and cross-modal interaction, SimBase achieves these results with a straightforward approach using temporal convolution and element-wise multiplication.
On the ActivityNet Captions dataset, SimBase attains similar performance to UnLoc at R@0.7 and shows a 1 point improvement at R@0.5.

\section{Conclusion}
This paper presents SimBase, a simple baseline for temporal video grounding.
Unlike existing works that devise complicate archetechures for temporal relationship modeling and multimodal fusion meachnism, this work answers the question: \textit{How effective can a simplified approach be}?
To explore this, we develop SimBase, a model that utilizes temporal convolutional layers instead of elaborate temporal structures.
For cross-modal interaction, SimBase relies solely on an element-wise product rather than intricate multimodal fusion techniques.
Notably, SimBase achieves state-of-the-art results on two large-scale datasets.
As a simple yet effective baseline, we hope SimBase inspires new ideas and simplifies the evaluation of future temporal video grounding model designs.

\bibliography{citations_SimBase.bib}
\bibliographystyle{IEEEtran}

\end{document}